%% file: main.tex
\documentclass[sigconf,authorversion]{acmart}
\usepackage{multirow}
\AtBeginDocument{%
  \providecommand\BibTeX{{%
    \normalfont B\kern-0.5em{\scshape i\kern-0.25em b}\kern-0.8em\TeX}}}

\copyrightyear{2023}
\acmYear{2023}
\acmDOI{}

\acmConference[CBMI '23]{20th International Conference on Content-based Multimedia Indexing}{September 20--22,
  2023}{Orleans, FR}
\acmPrice{}
\acmISBN{}

\begin{document}

\title[Spiking-FER: Event-based FER with SNNs]{Spiking-Fer: Spiking Neural Network for Facial Expression Recognition With Event Cameras}


\author{Sami Barchid}
\affiliation{%
  \institution{Univ. Lille, CNRS, Centrale Lille, UMR 9189 CRIStAL, Centre for Digital Systems}
  \city{F-59000 Lille}
  \country{France}
}
\email{sami.barchid@univ-lille.fr}

\author{Benjamin Allaert}
\affiliation{%
  \institution{IMT Nord Europe, Institut Mines-Télécom, Univ. Lille, Centre for Digital Systems}
  \city{F-59000 Lille}
  \country{France}
}
\email{benjamin.allaert@imt-nord-europe.fr}

\author{Amel Aissaoui}
\affiliation{%
  \institution{IMT Nord Europe, Institut Mines-Télécom, Univ. Lille, Centre for Digital Systems}
  \city{F-59000 Lille}
  \country{France}
}
\email{amel.aissaoui@imt-nord-europe.fr}

\author{José Mennesson}
\affiliation{%
  \institution{IMT Nord Europe, Institut Mines-Télécom, Univ. Lille, Centre for Digital Systems}
  \city{F-59000 Lille}
  \country{France}
}
\email{jose.mennessoni@imt-nord-europe.fr}

\author{Chaabane Djéraba}
\affiliation{%
  \institution{Univ. Lille, CNRS, Centrale Lille, UMR 9189 CRIStAL, Centre for Digital Systems}
  \city{F-59000 Lille}
  \country{France}
}
\email{chaabane.djeraba@univ-lille.fr}

\renewcommand{\shortauthors}{Barchid, et al.}

\begin{abstract}
  Facial Expression Recognition (FER) is an active research domain that has shown great progress recently, notably thanks to the use of large deep learning models. However, such approaches are particularly energy intensive, which makes their deployment difficult for edge devices. To address this issue, Spiking Neural Networks (SNNs) coupled with event cameras are a promising alternative, capable of processing sparse and asynchronous events with lower energy consumption. In this paper, we establish the first use of event cameras for FER, named "Event-based FER", and propose the first related benchmarks by converting popular video FER datasets to event streams. To deal with this new task, we propose "Spiking-FER", a deep convolutional SNN model, and compare it against a similar Artificial Neural Network (ANN). Experiments show that the proposed approach achieves comparable performance to the ANN architecture, while consuming less energy by orders of magnitude (up to 65.39x). In addition, an experimental study of various event-based data augmentation techniques is performed to provide insights into the efficient transformations specific to event-based FER.
\end{abstract}



\begin{teaserfigure}
  \centering
  \includegraphics[width=0.80\textwidth]{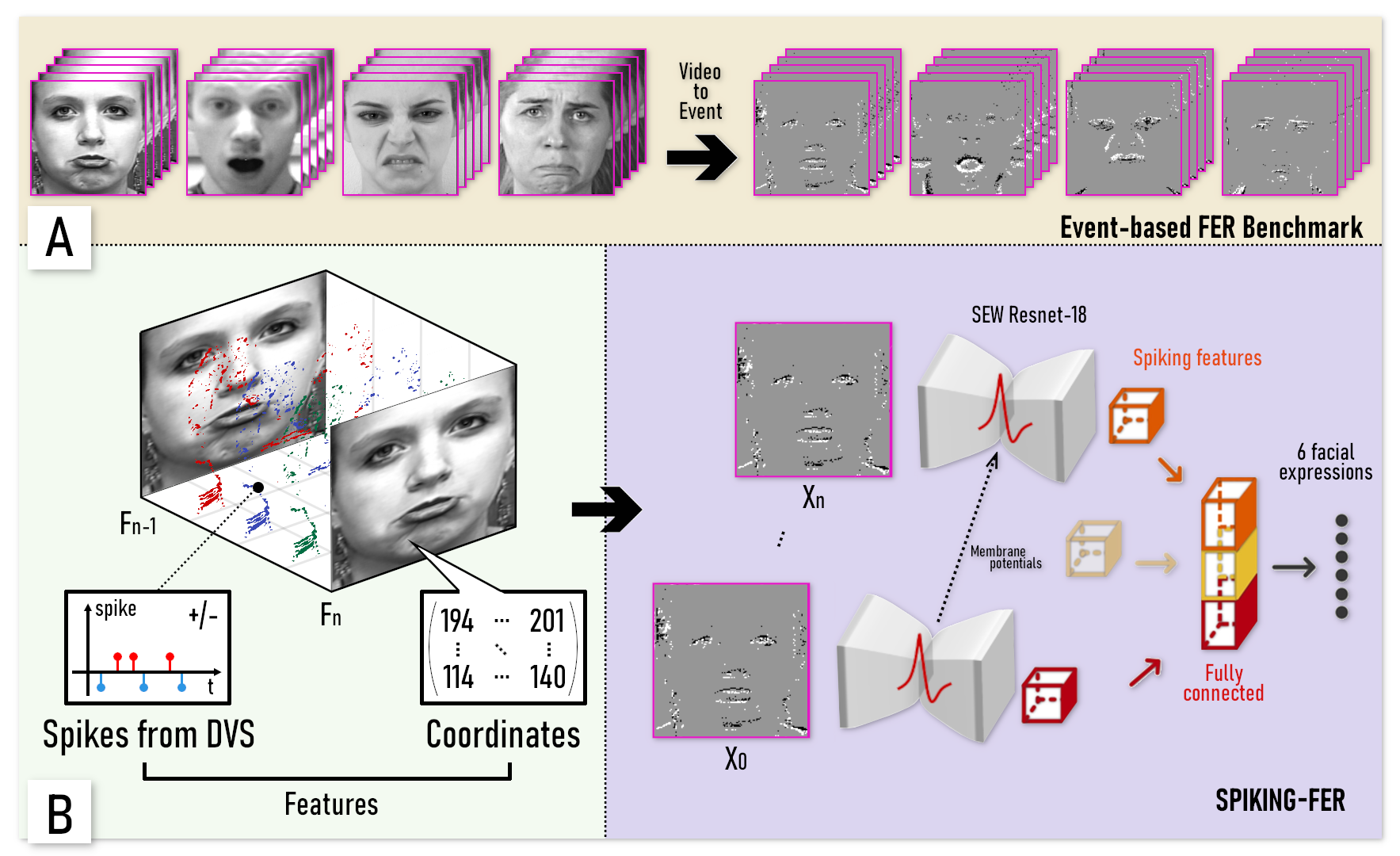}
  \caption{Overview of the proposed contributions to FER using event-based cameras and SNNs. A) A new benchmark based on event-driven data. B) An innovative Deep convolutional SNN model adapted for FER.}
  \label{fig:overview}
\end{teaserfigure}


\maketitle

\input{1_introduction.tex}
\input{2_related_works.tex}

\input{3_method.tex}

\input{4_1_setup.tex}
\input{4_experiments.tex}
\input{5_conclusion.tex}

\clearpage
\bibliographystyle{ACM-Reference-Format}
\bibliography{sample-base}


\end{document}

%% file: 1_introduction.tex
\section{Introduction}
\label{sec:introduction}

Facial Expression Recognition (FER) has received particular interest in recent years given its diverse and practical applications in computer vision, e.g., security, health, and communication. So far, efficient learning-based models have been proposed for FER \cite{li2020deep}. However, these methods often overcome the energy consumption constraint \cite{garcia2019estimation}. The convergence between the need to reinforce the complexity of learning models and the requirement of a physical platform at the cutting edge of technology induces heavy consequences in terms of energy consumption and has repercussions on the environment of the planet.

Spiking Neural Networks (SNNs) have become popular due to practical interest in terms of addressable complexity and energy efficiency than artificial neural networks (ANN). The neurons in SNNs asynchronously transmit information through sparse and binary spikes, enabling event-driven computing. Unlike ANN requiring dense matrix operations, energy consumption occurs only when a spike is generated on a neuromorphic chip. Hence, SNNs can significantly improve the energy efficiency of artificial intelligence systems.

\input{figures/intro_fig.tex}

Although SNNs are less energy-consuming, they do not match the performance of ANNs in computer vision. Inspired by the advances in ANNs, many recent methods are proposed to improve the performance of SNNs, e.g., ANN to SNN conversion \cite{conversion_snn}, innovative architectures \cite{spikfer}, new features encoding methods \cite{snn_objectdetection}, and data augmentation \cite{eventdrop}). However, very few studies have focused on FER, mainly due to the lack of training data.

In this paper, we propose an innovative framework for FER using SNNs, as illustrated in Fig. \ref{fig:overview}. First, a conversion framework using the V2E converter \cite{v2e} is proposed in order to preprocess the well-known FER video datasets and generate event-based streams which are a suitable format for SNN architecture. Then, the Spiking-FER model is trained using Surrogate Gradient Learning \cite{surrogate}, which enables the applicability of the backpropagation algorithm for deep SNN architectures. Finally, several experiments are conducted in order to evaluate the proposed model and compare it to a conventional ANN in terms of recognition accuracy and energy consumption.

Our proposal brings three novelties:
\begin{itemize}
    \item \textbf{Event-based FER Benchmark}: we provide a reproducible protocol to generate event-based FER benchmarks from the most popular video FER datasets which is, to the best of our knowledge, the first event-based benchmark proposed for FER.
    \item \textbf{New SNN Model Architecture}: we propose a new end-to-end deep convolutional SNN method, called "Spiking-FER" that encodes the event streams into spiking features that are then used by the output accumulator in order to predict the facial expression.
    \item \textbf{Event Data Augmentation for FER}: we analyze the performance of Spiking-FER and its corresponding ANN architecture depending on popular Event Data Augmentations (EDAs) to investigate their impacts on event-based FER.
\end{itemize}

The paper is structured as follows. In Section \ref{sec:related_works}, we review relevant recent methods for FER and on the evolution of the SNN for event-based vision. Section \ref{sec:event_based_fer} presents our SNN model architecture and the training process on event-based FER data. Section \ref{sec:experimental_setup} introduces the experimental setup including datasets and evaluation protocols. The experimental results are provided in Section \ref{sec:experiments}. Finally, we discuss the results and future work in Section \ref{sec:conclusion}.

%% file: figures/intro_fig.tex

%% file: 2_related_works.tex
\section{Related Works}
\label{sec:related_works}


\subsection{Facial Expression Recognition}
\label{subsec:fer}
FER methods could be classified into two categories: static (frame-based) and dynamic (sequence-based) methods. Frame-based methods extract spatial features from still images. They rely on hand-crafted features \cite{takalkar2020manifold} or learned features \cite{9039580} using mainly CNNs \cite{9039580}, but recently transformer architecture has also come into play \cite{aouayeb2021learning}. Sequence-based methods are performed in different ways: either by aggregating frames, e.g., onset-apex-offset \cite{yao2018texture} or onset-apex \cite{poux2021dynamic}, or by using consecutive frames in order to encode the temporal information. These methods use mainly deep architectures such as 3D CNN \cite{pan2019deep}, Recurrent Neural Networks (RNN) \cite{zhang2018spatial}, and Transformers \cite{liu2021expression}.

Recently, motion has come into play, and has proven to be effective in sequence-based FER \cite{9039580}. Moreover, it has been proposed to address different challenges for FER (occlusions \cite{poux2021dynamic}, intensity \cite{kumar2022three}), taking advantage of the fact that inter-individual variation in motion is better suited for FER than appearance features \cite{canedo2019facial}. Performance improvement is achieved by increasing the complexity of learning approaches, especially by taking into account spatio-temporal encoding. However, this improvement comes usually with the expense of energy consumption.

\subsection{Event-based Vision paired with Spiking Neural Networks}
\label{subsec:cv_snn}

Recently, learning algorithms adapted from backpropagation such as surrogate gradient learning \cite{surrogate} enable the training of deep SNN architecture by solving the non-differentiability issue of spiking neurons \cite{snntorch}. Such directly trained deep architectures are the first of several attempts capable of tackling event-based vision problems of similar complexity to those addressed by ANNs currently, such as object detection \cite{snn_objectdetection}, semantic segmentation \cite{snn_segmentation}, and object tracking \cite{snn_tracking}. Furthermore, these architectures start to adapt state-of-the-art techniques from ANNs (Vision Transformer \cite{spikfer}, spatio-temporal attention \cite{snn_attention}, ...) to operate with spiking neurons thanks to this new ability of gradient-based optimization. This recent direction of directly trained SNNs coupled with event cameras demonstrates impressive efficiency by being able to outperform ANNs \cite{sami_localization} and showing reduced energy consumption by orders of magnitude \cite{loihi_survey}.

%% file: 3_method.tex
\section{Methodology}
\label{sec:event_based_fer}


\subsection{Problem Formulation}
\label{subsec:formulation}

\input{figures/formulation.tex}

During a time interval $\Delta_{\mathcal{T}}$, an event camera with a $H \times W$ resolution produces a set of $N$ asynchronous events $\mathcal{E} = \{e_i\}_{i=1}^{N}$. Each event $e_i$ of the sequence can be formulated as a tuple of 4 values: $e_i = \{x_i, y_i, t_i, p_i\}$, where $(x_i, y_i)$ correspond to the pixel coordinates, $t_i$ is the timestamp, and $p_i \in \{1,-1\}$ is the sign of the polarity change.

As the asynchronous nature of events is not appropriate for many computer vision approaches \cite{event_based_vision}, a popular event representation method \cite{binary_event_frame} is to discretize a stream of events $\mathcal{E}$ into a sequence of $T$ binary event frames $\mathbf{X}_T \in \mathbb{B}^{T\times 2 \times H \times W} = \{X_t\}_{t=1}^{T}$. In this work, it is done by accumulating events during $T$ subsequent time intervals $\frac{\Delta_{\mathcal{T}}}{T}$ to create the sequence of binary frames and thus the final spike tensor $\mathbf{X}_T$.

Event-based FER can be defined as follows: given an event sequence $\mathcal{E}$ obtained from capturing a subject that performs a facial expression, the objective is to recognize this expression as the appropriate label $c$ among $\mathcal{C}$ different classes. To do so, a model $f_\alpha(\cdot)$ with a set of learnable parameters $\alpha$ is trained such that: $c = f_\alpha(\mathbf{X}_T)$. The top of Fig. \ref{fig:problem_formulation} illustrates the formulation of event-based FER with the related notations. 

\subsection{Spiking-FER}
\label{subsec:spikfer}

Spiking-FER is represented by the model $f_\alpha(\cdot)$, where $\alpha$ denotes its synaptic weights. The bottom of Fig. \ref{fig:problem_formulation} illustrates an overview of the proposed Spiking-FER architecture.

\subsubsection{Spiking Neuron Model}
The proposed convolutional SNN architecture uses the Integrate-and-Fire (IF) neuron \cite{IFneuron} as the spiking neuron model. It accumulates input spikes weighted by the synaptic weights into a "membrane potential". When this membrane potential exceeds a certain threshold value, the neuron emits an output spike and is reset to zero. The discretized dynamics of a layer $l$ of IF neurons from Spiking-FER at a certain time-step $1 \le t \le T$ is described as follows:
\begin{align}
\label{equation_if_neuron}
U_t^l & = U_{t-1}^l + \mathcal{W}^{l} X^{l-1}_{t-1} - \theta X_t^l \\
\label{equation_if_neuron_2}
X_t^l & = \Theta(U_t^l - \theta) 
\end{align} where $U_t^l$ denotes the membrane potentials of the IF neurons, $\mathcal{W}^l$ is the set of synaptic weights, $X_t^l \in \mathbb{B}$ denotes the output spike tensor. $X_t^l$ consists of 1's when the related element of $U_t^l$ exceeds the threshold value $\theta$, and 0's otherwise. For simplicity, the threshold is set to $1$ for all layers (i.e., $\theta = 1$). This mechanism, formulated in Eq. \ref{equation_if_neuron_2} is known as the Heaviside step function ($\Theta(\cdot)$).

\subsubsection{Direct Training via Surrogate Gradient}
Spiking-FER is trained using Surrogate Gradient Learning \cite{surrogate,snntorch}, a popular and effective training approach for deep SNN models. An SNN can be expressed as a Recurrent Neural Network where the membrane potentials are internal states. Consequently, the synaptic weights can be trained using Backpropagation Through Time \cite{bptt}. The main issue is related to the backward pass, where $\Theta(\cdot)$ is not differentiable - i.e., its derivative is 0 almost everywhere, and $+\infty$ at 0 - causing the gradient chain to break ("dead neuron problem" \cite{snntorch}). Therefore, surrogate gradient learning solves this problem by employing the derivative of a continuous surrogate function $\sigma(\cdot)$ on the backward pass as an approximation of the derivative of $\Theta(\cdot)$. In Spiking-FER, we define $\sigma(x) = \frac{1}{\pi} \arctan(\pi x) + \frac{1}{2}$.

\subsubsection{Model Architecture}
Strongly related to \cite{sami_localization}, Spiking-FER consists of two modules: \textbf{(1)} a deep convolutional SNN encoder that encodes the event streams into spiking features; and \textbf{(2)} an output accumulator module \cite{snn_segmentation} that predicts the emotion of the sample from the encoded spiking features. 

The encoder is a SEW-ResNet-18 \cite{sew_resnet} architecture that outputs spiking feature vectors $F_t \in \mathbb{B}^d$, where $d$ is the number of output channels (in SEW-ResNet-18, $d=512$). At each time-step, these extracted spiking features are fed into the output accumulator module responsible for making the final prediction.

As shown in the rightmost part of Fig. \ref{fig:problem_formulation}, the output accumulator module is composed of one fully connected layer of artificial neurons and one linear classifier. Firstly, it accumulates the spiking features from all time-steps to obtain a single feature vector $\mathcal{F} \in \mathbb{R}^d$ such that:

\begin{equation}
    \mathcal{F} = \sum_{t=1}^{T} \mathcal{W} \times F_t
\end{equation}, where $\mathcal{W} \in \mathbb{R}^{d \times d}$ is the set of trainable weights in the fully connected layer. Then, the features $\mathcal{F}$ are fed into the linear classifier to obtain the final classification prediction.

The whole network is trained end-to-end using the cross-entropy loss.

%% file: figures/formulation.tex
\begin{figure*}[t]
\centering
\includegraphics[width=\textwidth]{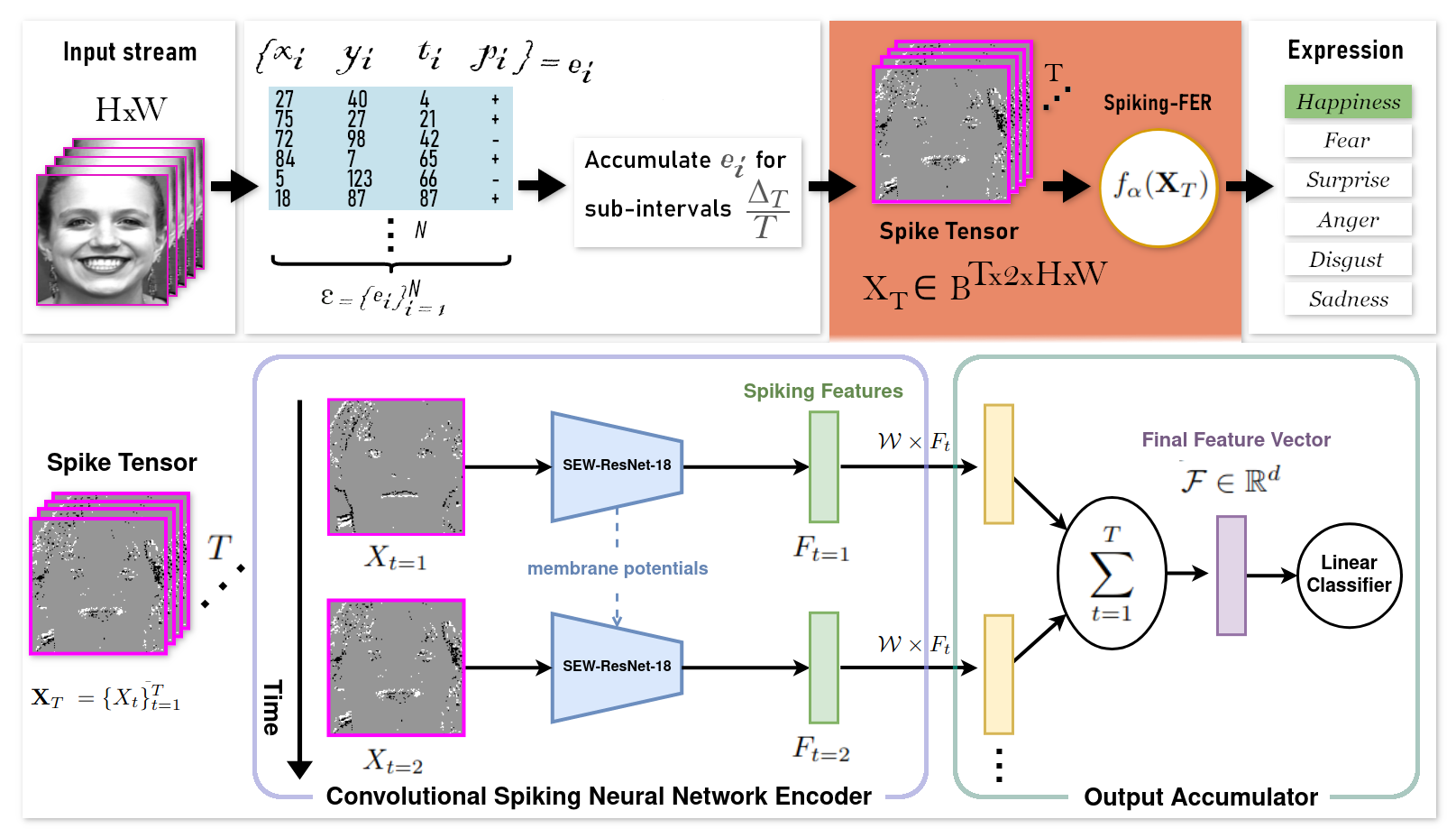}
\caption{Overview of the proposed framework. \textit{Top)} Formulation of Event-based Facial Expression Recognition. \textit{Bottom)} The Spiking-FER architecture where the convolutional SNN encoder is expressed as a recurrent neural network.}
\label{fig:problem_formulation}
\end{figure*}

%% file: 4_1_setup.tex
\section{Experimental Setup}
\label{sec:experimental_setup}

In this section, we present the experimental setup including datasets, evaluation protocols and the models configurations. 

\subsection{Video-to-Events Conversion}

To validate the applicability of event-based data and SNNs to FER applications, while being comparable to standard FER baselines \cite{fer_survey}, we convert some of the most popular video FER datasets: ADFES \cite{ADFES}, CASIA \cite{CASIA}, CK+ \cite{CKPlus}, and MMI \cite{MMI} to an event-based format. Each video of a given FER dataset is processed by two successive steps. The first step is a standardization of all frames \cite{benjamin_comparative_study}: the face of the represented subject is cropped and rotated based on 68 facial landmarks and converted to grayscale. Then, the resulting frame is resized to a resolution of $(200\times200)$. The second step corresponds to the conversion of the standardized video into events using v2e \cite{v2e}, a video-to-event converter for realistic events simulation, as illustrated in Fig. \ref{fig:bdd}. The code and parameters to reproduce the benchmark are available\footnote{The code will be released upon acceptance} .

\subsection{Evaluation Protocol}
\label{subsubsec:evaluation_protocol}
Models that are evaluated on an event-based FER dataset follow a $10$-fold cross-validation configuration: the given dataset is randomly split into 10 folds of equal size. The employed metric for every iteration is the top-1 classification accuracy. Finally, we report the mean accuracy score of the 10 folds.

\subsection{Implementation Details}
\label{subsec:implementation_details}
The experiments are implemented in PyTorch, Tonic \cite{tonic} and SpikingJelly \cite{spikingjelly} as our SNN simulator, and run on one NVIDIA A40 GPU. We train our models (Spiking-FER and ResNet-18) during 500 epochs, using an SGD optimizer, with a learning rate of $0.01$ and a cosine annealing scheduler \cite{cosine_annealing}, and keep the best performance on the validation set. A low-latency regime is adopted for Spiking-FER, with $T=6$.

\subsection{Comparison with ANN}
Since the convolutional SNN encoder of Spiking-FER is a SEW-ResNet-18, we choose a ResNet-18 \cite{resnet} model as the corresponding ANN. Similarly to the ANN model defined in \cite{bina_rep}, the spike tensor $\mathbf{X}_T$ is fed into the 2D-CNN by concatenating all binary event frames together along the time axis.

\input{figures/bdd.tex}

%% file: figures/bdd.tex
\begin{figure}[t]
\centering
\includegraphics[width=\columnwidth]{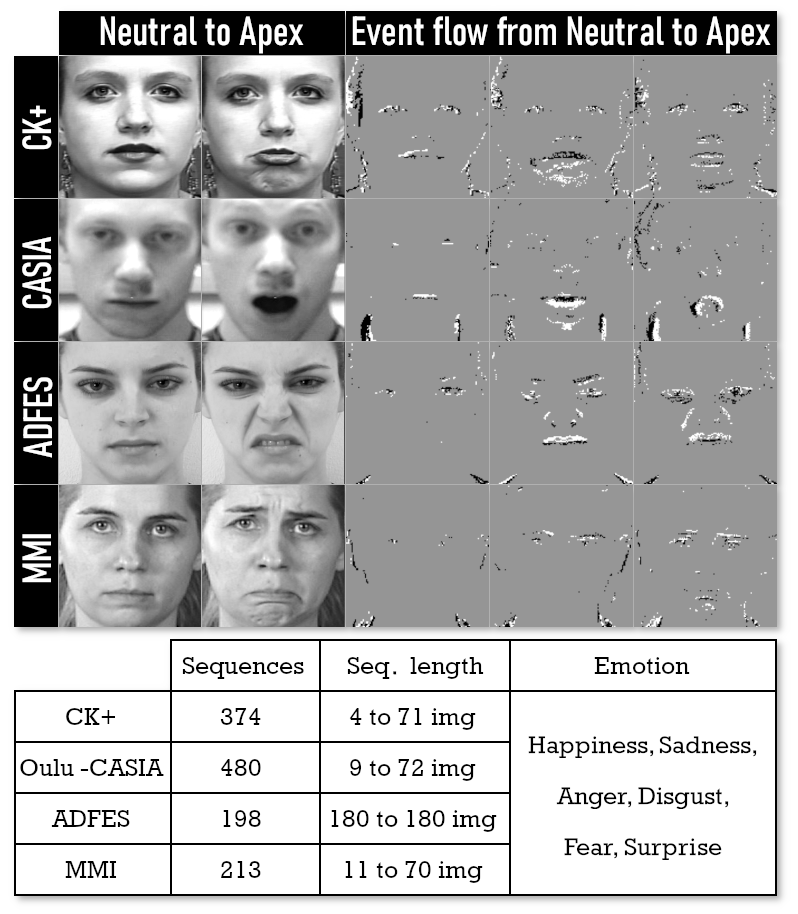}
\caption{Illustration of the proposed event-based FER benchmark. The video sequences are converted into events corresponding to the output of event cameras.}
\label{fig:bdd}
\end{figure}

%% file: 4_experiments.tex
\section{Experiments}
\label{sec:experiments}


\subsection{Study on Event Data Augmentation}
To investigate the impacts of popular event data augmentations (EDAs)\cite{nda}, given in Table \ref{tab:eda_summary}, on the model performance, the experiments are conducted in 2 successive parts: \textbf{(1)} an analysis of common EDAs; \textbf{(2)} an analysis on specific EDAs for either regularization of training with scarce datasets, or FER-specific transformation.

\input{tables/eda_summary.tex}

\subsubsection{Common EDAs}
Since EDAs can be applied in combination, the main objective of this part is to assess which EDA has the best impact when they are combined with each other. Therefore, we run all possible combinations of common EDAs, which gives a total of 32 experiments for a given dataset, as illustrated in Fig. \ref{fig:exp1}. 

\input{figures/expe1.tex}

The baseline results show that the SNN model performs better than the ANN model without augmentation - i.e., using only the original data from the event stream. Often observed in neural network training, data augmentation tends to significantly improve performance. This is especially true for FER, where databases are scarce. On ANNs, we observe that all the EDAs combinations have a positive or null impact, unlike the SNNs, where some EDAs combinations tend to decrease the performances. Among the EDA methods, the combination \{\textit{Crop}, \textit{H Flip} and \textit{Noise}\} significantly improves the performance on both ANNs and SNNs, except for the MMI dataset, where the improvement is less significant. This can be explained by the greater complexity of the data, where greater head pose variations and variety of facial movement patterns appear.

Then, we evaluate the accuracy scores of all folds for all experiments, which gives 320 scores. We perform a multivariate regression analysis on this population of 320 scores by considering the applied EDAs as categorical independent variables. For a given EDA, the regression analysis gives an approximation of the expected benefit in performance. Fig. \ref{fig:exp2} shows the results of the regression analysis for each dataset. According to the regression coefficients, $Crop$ and $HFlip$ have generally a positive impact, which suggest that they are well adapted for event-based FER. These methods cover well the small variations, e.g., face translation or image resolution changes, on the data observed in the different databases that compose the benchmark. However, $Reverse$ that reports either non-significant results or negative impacts in all cases. This can be explained by the fact that the activation of a facial expression follows a temporal sequence induced by the facial muscles. In this case, the reversal of the event flow is not consistent, especially since in this benchmark, where the sequences only go from the neutral state to the apex. $PolFlip$ highlights the differences between Spiking-FER and the ANN: while Spiking-FER constantly reports negative effects, the ANN model obtains a positive impact. This suggests that SNNs do not benefit from $PolFlip$ for event-based FER.

\input{figures/expe2.tex}

\subsubsection{Specific EDAs}
We keep the best-performing combinations of common EDAs and evaluate the specific ones. For a given dataset, the best combination of common EDAs is defined as the highest mean accuracy score obtained on the $10$-fold cross-validation. Fig. \ref{fig:exp3} reports the results obtained with and without these specific EDAs, adapted to the event flows for FER. Considering the performances, we note that the combination of \textit{EventDrop}, which regularizes the training of neural networks on limited datasets, and \textit{Mirror}, which transforms the visual aspect of a subject's face, is perfectly adapted to augment facial expressions for both ANNs and SNNs. In addition, to improve the performance of the models, the performance gap between the ANNs and SNNs models is significantly reduced, especially for the ADFES dataset. Both EDAs have been designed to adapt to inter-individual variation, e.g., face symmetry and expression activation time.

\input{figures/expe_3.tex}

\subsection{Estimation of Energy Consumption}
Similarly to \cite{snn_segmentation,sami_localization}, we compare the energy efficiency of Spiking-FER and a similar ANN when simulated on a 45nm CMOS chip \cite{cmos}.

The estimation methodology is described as follows: firstly, we quantify the spiking rate of each layer, as spiking neurons consume energy only when generating a spike. The spiking rate of a given layer $l$ is calculated as follows:
\begin{equation}
Rs(l) = \frac{\text{\# spikes of } l \text{ over all time-steps}}{\text{\# neurons of } l}
\end{equation}

Secondly, we compute the total floating-point operations (FLOPs) of a layer of spiking neurons ($FLOPs_{SNN}$) by using the FLOPs of the same layer in a non-spiking neural network ($FLOPs_{ANN}$) and the spike rate of the spiking neuron layer:

\begin{align}
\label{eq:flops_snn}
FLOPs_{SNN}(l) & = FLOPs_{ANN}(l) \times Rs(l) \\
\label{eq:flops_ann}
FLOPs_{ANN}(l) & = 
	\begin{cases}
	k^2 \times O^2 \times C_{in} \times C_{out} & \text{if } l \text{ is Conv.} \\
    C_{in} \times C_{out} & \text{if } l \text{ is Linear.}
	\end{cases}
\end{align} In Equation \ref{eq:flops_ann}, $k$ represents the kernel size, $O$ represents the size of output feature maps, $C_{in}$ represents the number of input channels, and $C_{out}$ represents the number of output channels.

Finally, the total energy consumption of a model can be estimated on CMOS technology \cite{cmos} by using the total FLOPs across all layers. Table \ref{tab:energy_operations} presents the energy cost of relevant operations in a 45nm CMOS process. MAC operation in ANNs requires one addition (32bit FP ADD) and one FP multiplication (32bit FP MULT) \cite{mac_ann}, whereas SNNs require only one FP addition per MAC operation due to binary spike processing. The total energy consumption of ANNs and SNNs are represented by $E_{ANN}$ and $E_{SNN}$, respectively.

\begin{align}
E_{ANN} & = \sum\limits_{l}{FLOPs_{ANN}(l)} \times E_{MAC} \\
E_{SNN} & = \sum\limits_{l}{FLOPs_{SNN}(l)} \times E_{AC}
\end{align}

\begin{table}[t]
\centering
\caption{Energy table for a 45nm CMOS process (from \cite{snn_segmentation}).}
\label{tab:energy_operations}
\begin{tabular}{cc}
\textbf{Operation} & \textbf{Energy} (pJ) \\ \hline
32bit FP MULT  ($E_{MULT}$)     & 3.7                  \\
32bit FP ADD  ($E_{ADD}$)     & 0.9                  \\
32bit FP MAC  ($E_{MAC}$)     & 4.6   ($= E_{MULT} + E_{ADD}$)             \\
32bit FP AC    ($E_{AC}$)    & 0.9                 
\end{tabular}
\end{table}

Table \ref{tab:energy_efficiency} reports the mean inference energy estimation for each dataset. Similarly to previous works \cite{sami_localization}, Spiking-FER shows better energy efficiency by orders of magnitude (from $47.42\times$ to $65.39\times$ more efficient), which proves the applicability of SNNs for low-power FER application on edge devices.

\input{tables/energy_efficiency.tex}

%% file: tables/eda_summary.tex
\begin{table}[t]
\fontsize{7}{12}\selectfont
\centering
\caption{Summary of EDAs. Common EDAs and Specific EDAs are respectively in italic and in bold.} 
\label{tab:eda_summary}
\resizebox{\columnwidth}{!}{%
\begin{tabular}{ll}
\hline
\textbf{EDA}                                             & \textbf{Description}                                    \\ \hline
\begin{tabular}[c]{@{}c@{}}$Crop$\end{tabular} &
  \begin{tabular}[c]{@{}c@{}}Spatial crop of the whole sequence with a random scale\end{tabular} \\ 
\begin{tabular}[c]{@{}c@{}}$HFlip$\end{tabular}   & Horizontal flip of the whole sequence                   \\ 
\begin{tabular}[c]{@{}c@{}}$Noise$ ($BA$)\end{tabular} &
  \begin{tabular}[c]{@{}c@{}}Noisy events due to corrupted pixels in event cameras \cite{ba}. \end{tabular} \\ 
\begin{tabular}[c]{@{}c@{}}$PolFlip$\end{tabular} & Flip of polarity (i.e., $p_i = - p_i$ for all events)   \\ 
\begin{tabular}[c]{@{}c@{}}$Reverse$\end{tabular} &
  \begin{tabular}[c]{@{}c@{}}Reverse the orders of events.\end{tabular} \\ 
\textbf{EventDrop} \cite{eventdrop}                                          & Randomly drops events spatially, temporally or globally \\ 

\textbf{Mirror}                                                             & Mirrors the left or right half of the sequence \\   \hline     
\end{tabular}%
}
\end{table}

%% file: figures/expe1.tex
\begin{figure*}[t]
\centering
\includegraphics[width=\textwidth,height=4cm]{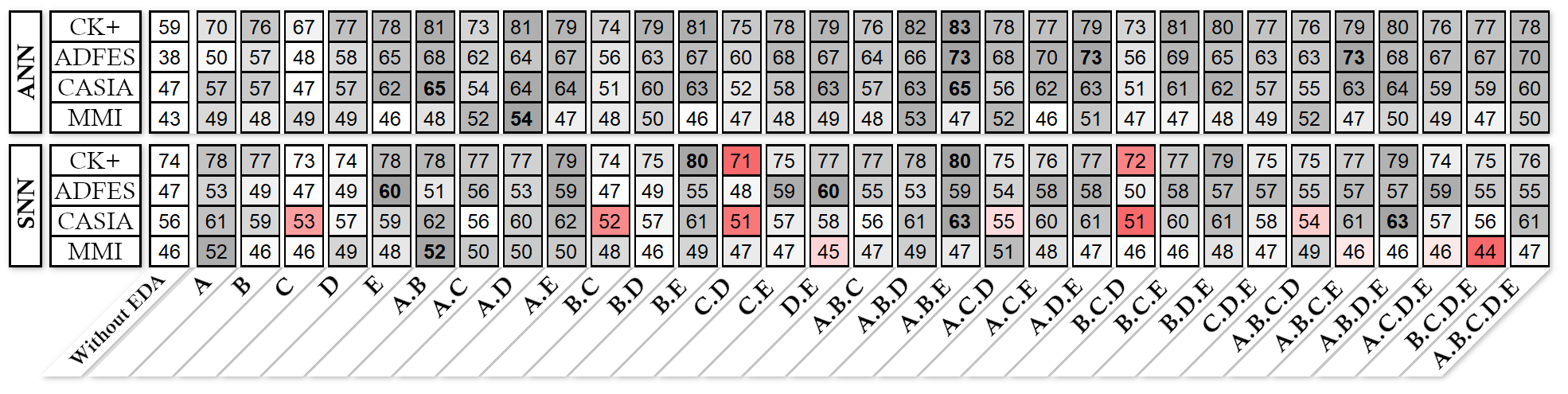}
\caption{Acc. obtained according to combinations of common EDA; (A) \textit{H Flip}; (B) \textit{Noise}; (C) \textit{Reverse}; (D) \textit{Pol Flip}; (E) \textit{Crop}.}
\label{fig:exp1}
\end{figure*}

%% file: figures/expe2.tex
\begin{figure}[t]
\centering
\includegraphics[width=0.6\columnwidth]{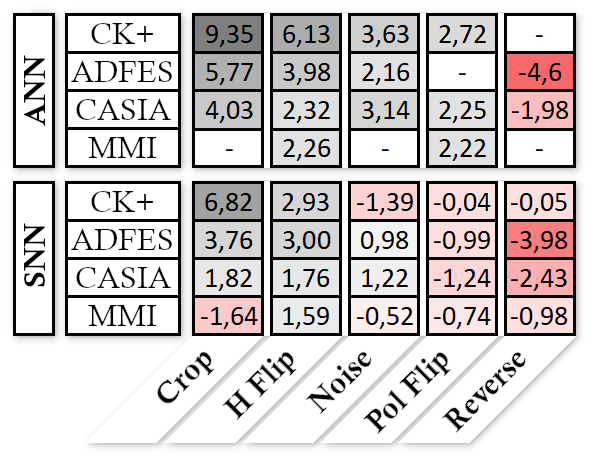}
\caption{Significative regression coefficients (p-value $< 0.05$) calculated on the 320 scores, corresponding to each common EDA for the different datasets (higher is better).}
\label{fig:exp2}
\end{figure}

%% file: figures/expe_3.tex
\begin{figure}
\centering
\includegraphics[width=0.7\columnwidth]{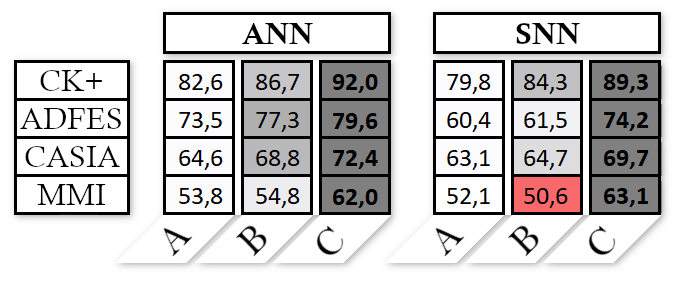}
\caption{Comparison of the performances obtained with and without the specific EDAs. (A) best configuration based on common EDA; (B) with the addition of \textit{EventDrop}; (C) with the addition of \textit{EventDrop} and \textit{Mirror}.}
\label{fig:exp3}
\end{figure}

%% file: tables/energy_efficiency.tex
\begin{table}
\centering
\caption{Mean estimated inference energy of Spiking-FER ($E_{SNN}$) and the similar ANN architecture ($E_{ANN}$).}
\label{tab:energy_efficiency}
\resizebox{\columnwidth}{!}{%
\begin{tabular}{c|ccc}
                   & $E_{ANN}$ (mJ)           & $E_{SNN}$ (mJ) & $E_{ANN}/E_{SNN}$ \\ \cline{2-4} 
\textbf{DVS-ADFES} & \multirow{4}{*}{1428.67} & 21.85          & 65.39             \\
\textbf{DVS-CASIA} &                          & 26.22          & 54.49             \\
\textbf{DVS-CK+}   &                          & 27.15          & 52.62             \\
\textbf{DVS-MMI}   &                          & 30.13          & 47.42            
\end{tabular}%
}
\end{table}

%% file: 5_conclusion.tex
\section{Conclusion }
\label{sec:conclusion}

In this work, we introduced \textit{event-based benchmarks for Facial Expression Recognition} (FER) and proposed a new SNN architecture named \textit{Spiking-FER}. We applied traditional augmentation techniques adapted to event streams, along with two specific techniques - \textit{EventDrop} \cite{eventdrop} and \textit{Mirror} - that led to significant improvements in our model's performance. Our proposed approach achieved similar performance to a traditional Artificial Neural Network (ANN) while consuming much less energy (up to 65.39×). Our future work will extend this study to other applications such as gesture or action analysis.